\documentclass[10 pt, conference]{IEEEtran}  

\IEEEoverridecommandlockouts                              

\usepackage{graphics} 
\usepackage{epsfig} 
\usepackage{mathptmx} 
\usepackage{times} 
\usepackage{amsmath} 
\usepackage{amssymb}  
\usepackage{amsthm}  
\usepackage{enumerate} 
\usepackage{subfig}
\usepackage{makecell} 
\usepackage{multirow}
\usepackage{mathtools}
\usepackage[english, onelanguage,ruled,vlined, longend, titlenotnumbered]{algorithm2e}
\usepackage{enumitem}
\usepackage{bbold}

\newenvironment{sizeddisplay}[1]
 {\par\nopagebreak#1\noindent\ignorespaces}
 {\nopagebreak\ignorespacesafterend}

\newtheorem{definition}{Definition}

\newtheorem{thm}{Theorem}
\newtheorem{prop}{Proposition}

\newtheorem{expl}{Example}

\newcommand{\bfa}{\mathbf{a}}

\newcommand{\hbfa}{\hat{\bfa}}
\newcommand{\bfA}{\mathbf{A}}
\newcommand{\bfb}{\mathbf{b}}
\newcommand{\hbfb}{\hat{\mathbf{b}}}

\newcommand{\bfC}{\mathbf{C}}

\newcommand{\bfD}{\mathbf{D}}

\newcommand{\bfF}{\mathbf{F}}

\newcommand{\bfg}{\mathbf{g}}

\newcommand{\bfI}{\mathbf{I}}

\newcommand{\bfM}{\mathbf{M}}

\newcommand{\bfn}{\mathbf{n}}

\newcommand{\bfp}{\mathbf{p}}

\newcommand{\bfP}{\mathbf{P}}

\newcommand{\bfQ}{\mathbf{Q}}

\newcommand{\bfR}{\mathbf{R}}

\newcommand{\hbfR}{\hat{\bfR}}

\newcommand{\bfu}{\mathbf{u}}

\newcommand{\bfv}{\mathbf{v}}

\newcommand{\hbfv}{\hat{\bfv}}

\newcommand{\bfw}{\mathbf{w}}

\newcommand{\bfx}{\mathbf{x}}

\newcommand{\hbfx}{\hat{\mathbf{x}}}

\newcommand{\bfy}{\mathbf{y}}

\newcommand{\RR}{{\mathbb R}}

\newcommand{\hchi}{\hat{\chi}}

\newcommand{\boUpsilon}{\boldsymbol{\Upsilon}}
\newcommand{\bfzero}{\mathbf{0}}

\DeclareMathOperator*{\argmax}{argmax}

\DeclareMathAlphabet{\mathsfbfit}{OT1}{lmss}{bx}{sl}

\newcommand{\fixX}{x}
\newcommand{\bodyX}{\text{\sc x}}
\newcommand{\body}[1]{\text{\sc #1}}
\newcommand{\fixV}{V}
\newcommand{\bodyV}{B}

\newcommand{\fixF}{\bfA}
\newcommand{\fixd}{d}
\newcommand{\fixC}{\bfC}
\newcommand{\fixu}{u}
\newcommand{\bodyF}{\Phi}
\newcommand{\bodyd}{\body{d}}
\newcommand{\bodyC}{\Gamma}
\newcommand{\bodyu}{\body{u}}

\newcommand{\elem}[1]{#1}
\newcommand{\action}{*}

\newcommand{\grouplaw}{ } 

\newcommand{\biggrouplaw}{\bullet}

\usepackage{siunitx}
\usepackage{pgfplots,pgfplotstable}
\usepackage{tikz}
\usetikzlibrary[patterns]
\pgfplotsset{compat=1.12}
\usetikzlibrary{plotmarks}
\usepackage[nospace]{cite}
\usetikzlibrary{arrows,shapes,positioning}
\usetikzlibrary{calc,trees,positioning,arrows,chains,shapes.geometric,%
	decorations.pathreplacing,decorations.pathmorphing,shapes,%
	matrix,shapes.symbols}
\usetikzlibrary{decorations.markings}

\usepackage[top=54pt, bottom=54pt, left=54pt,
      right=54pt]{geometry}

\usepackage{setspace}

\usepackage{hyperref}
\hypersetup{hidelinks}

\author{Paul Chauchat$^{1}$,   Axel Barrau$^{2 }$, and Silv\`ere Bonnabel$^{3}$
	\thanks{$^{1}$Aix-Marseille Univ, CNRS, LIS, Marseille, France
	{\tt\small paul.chauchat@lis-lab.fr}}%
 \thanks{$^{2}$OFFROAD, 5 rue Charles de Gaulle, Alfortville, France
			\thanks{$^{3}$Centre for Robotics, MINES Paris, PSL Research University,  60 Boulevard Saint-Michel, 75006 Paris, France
		{\tt\small silvere.bonnabel@mines-paristech.fr}}%
	}%
}

\begin{document}
\title{\LARGE \bf Invariant Smoothing  for Localization: Including the IMU Biases}
\author{ Paul Chauchat,  Axel Barrau, Silvère Bonnabel 
    \thanks{P.~Chauchat is with Aix-Marseille Univ, CNRS, LIS, France
	({\tt\small paul.chauchat@lis-lab.fr}).  A~Barrau is with OFFROAD. S.~Bonnabel is with MINES Paris PSL, PSL Research University, France ({\tt\small silvere.bonnabel@mines-paristech.fr}). }
}

\maketitle
\begin{abstract}
In this article we investigate smoothing (i.e., optimisation-based)  state estimation techniques for robot localization using an IMU aided by other localization sensors. We more particularly focus on Invariant Smoothing (IS), a variant based on the use of   Lie groups within the invariant filtering framework. We bring the recently introduced Two Frames Group (TFG) to bear on the problem of Invariant Smoothing, to better take into account the IMU biases, as compared to the state-of-the-art in localization and navigation.  Experiments based on the KITTI dataset show  the proposed framework compares favorably to the state-of-the-art smoothing methods in terms of robustness in some  challenging situations.
\end{abstract}
\section{INTRODUCTION}
Lie group embeddings have become standard tools in navigation and mobile robotics over the last decade, see e.g.,  \cite{chirikjian2011stochastic2,  sola2018micro, BayesianLieGroups2011}.  Extended Kalman filters or observers based on the classical groups  $SE(3)$ and $SE(2)$ have been introduced and successfully used, see \cite{hua2014implementation,bourmaud2013discrete,bonnabel2008symmetry}. Novel Lie groups were then introduced   and proved to come with similar properties to accomodate more complex dynamics, see \cite{barrau2017invariant,barrau2015non}. 
They came with  a new class of estimators, under the name of Invariant Extended Kalman Filter (IEKF)   \cite{barrau2017invariant, barrau2018annual}. The theoretical properties include  convergence guarantees \cite{barrau2017invariant},  consistency properties  \cite{barrau2018annual}, and have led to applications in various fields,   \cite{barrau2018annual, van2020invariant, hartley2020contact, pavlasek2021invariant, wu2017invariant, caruso2019magneto, heo2018consistent, mahony2017geometric} and in the industry \cite{barrau2018annual}.  Observer design also benefited from the introduction of those new groups, see e.g., \cite{wang2020hybrid, hashim2021gps}, and they are also studied in their own right \cite{van2021autonomous}.

The success of the IEKF for models from inertial navigation relies on the introduction of the Lie group of double spatial direct isometries $SE_2(3)$, or extended poses, in \cite{barrau2017invariant,barrau2015non}, and  \cite{brossard2021associating} for a more thorough exposition. However, inertial measurement unit (IMU) biases could not be properly included into this Lie group structure, although they are unknown and need be estimated. Hence, they were treated as additional parameters leading to an ``imperfect'' invariant framework \cite{barrau2015non, van2020invariant}. However, the recently introduced  Two Frame Group (TFG) structure \cite{barrau2022geometry}   additionally allows properly including the accelerometer bias.

If we turn to state estimation, smoothing has become the most popular technique in robotics for simultaneous localisation and mapping (SLAM) and visual odometry, thanks to its reducing the consequences of linearisation errors \cite{forster2016preintegration}. It is now often used in inertial navigation too \cite{indelman2013incremental, walsli2018invariant}. Leveraging the framework of Invariant filtering for smoothing, a new estimation algorithm was recently proposed, namely Invariant Smoothing (IS)  \cite{chauchat2018invariant}, see also \cite{walsli2018invariant}, which delivers ``physically consistent'' estimates  \cite{chauchat2022smoothing}, although no convergence guarantee was derived. While recent SLAM and attitude estimation methods with convergence guarantees were proposed, they do not consider biased acceleration measurements \cite{hashemi2022global,yi2021almost,dellaert2020shonan}.

This paper provides all the tools required to use the TFG structure in a new framework, namely smoothing, in order to tackle localization using biased IMUs. It completes \cite{barrau2022geometry}, providing formulas for the adjoint map in particular.  To evaluate the proposed framework, experiments were conducted based on data from the KITTI dataset \cite{geiger2013kitti}, focusing in particular on the difficult problem of   localization  with no prior information, known as ``in-flight alignment" problem in inertial navigation when the sensors used are the IMU and GNSS \cite{wu2013integration,   ouyang2022optimization,chang2021strapdown}.

The paper is organized as follows. The main principles of Invariant Filtering and Smoothing   are recalled in Section~\ref{sec:lie_group}. The considered application to localization is presented in Section~\ref{sec3}, as well as the presentation of the TFG and its properties for the considered problem. The proposed invariant smoother is detailed in Section~\ref{expl::nav_IMU}, and the differences to the existing state-of-the-art smoothers are explained. Experimental results  and comparisons are presented in Section~\ref{sec:expe} and show the proposed IS based on the TFG  favorably compares to state-of-the-art  smoothing schemes \cite{forster2016preintegration, dellaert2016new}, and to the former ``imperfect'' IS based on $SE_2(3)$.

	\section{Smoothing on Lie groups}\label{sec:lie_group}

	We first briefly recall the invariant filtering framework \cite{barrau2017invariant, barrau2018annual}. The reader is referred to \cite{ sola2018micro} for a general presentation. We consider a state $\chi \in G$, with $G$ a  Lie group of dimension $q$. Its Lie algebra $\mathfrak{g}$ is identified with $\RR^q$. Thus we consider its exponential map to be defined as $\exp : \RR^q \rightarrow G$. We denote its local inverse by $\log$. We recall the notion of adjoint operator matrix of $\chi \in G$, $\mathbf{Ad}_{\chi}$, which satisfies
	\begin{equation}
	\forall \chi \in G, \xi \in \RR^q,\ \chi^{-1} \exp(\xi) \chi =  \exp(\mathbf{Ad}_{\chi} \xi) 
	\label{eq:adjoint}
	\end{equation}
	Automorphisms are bijective maps $\phi : G \rightarrow G$ satisfying \begin{equation}
	    \phi(\chi \eta) = \phi(\chi) \phi(\eta)\quad\text{for}~\chi, \eta \in G.\label{auto:prop}
	\end{equation}The Lie group Lie algebra   correspondance, see  \cite{barrau2019linear}, ensures for $\emph{any}$ automorphism $\phi$ there is $\mathbf{M} \in \RR^{q\times q}$ so that
	\begin{equation}
	\forall (\chi, \xi) \in G\times \RR^q,\ \phi(\chi \exp(\xi)) = \phi(\chi) \exp(\mathbf{M} \xi),
	\label{eq:LGLA}
	\end{equation}which is closely related to  the \emph{log-linearity} property of \cite{barrau2017invariant}. 
The operator $\nu \mapsto  \chi^{-1} \nu \chi $ is easily checked to be a group automorphism, and  we see indeed from \eqref{eq:adjoint} that   $\mathbf{M} =\mathbf{Ad}_{\chi}$. We define random variables on Lie groups through the exponential, following  \cite{chirikjian2011stochastic2,bourmaud2013discrete, brossard2021associating,barrau2018annual}. The probability distribution $\chi \sim \mathcal{N}_L(\bar{\chi},\mathbf{P})$ for the random variable $\chi \in G$ is defined as 
\begin{equation}
	\chi =  \bar{\chi} \exp \left(\xi\right), \text{~} \xi \sim \mathcal{N}\left(\mathbf{0}, \mathbf{P}\right), \label{eq:left_distrib}
\end{equation}
where $\bar{\chi} \in G$, and $\bfP$ is a covariance matrix. 	In the following, we   consider a discrete-time  trajectory $(\chi_i)_{0\leq i \leq n}$, denoted $(\chi_i)_i$ for brevity, of the following system
 	\begin{equation}
		\chi_0 \sim \mathcal{N}_L(\bar{\chi}, \mathbf{P}_0), \quad \chi_{i+1} = f_i(\chi_i) \exp(\bfw_i), \quad
		y_k = h_k(\chi_{I_k}) + \bfn_k 
		\label{eq:l-is_system}
	\end{equation}
 where $f_i$, $h_k$ are the dynamics and observation functions respectively, $\mathbf{P}_0 \in \RR^{q \times q}$ the initial state error covariance, $\bfw_i, \bfn_k$ are white noises of covariance $\bfQ_i$ and $\mathbf{N}_k$, and $\chi_{I_k}$ denotes a subset of the states which are involved in the measurements at $t_k$.

 \subsection{Smoothing on Lie groups}
	We first briefly recall the Invariant Smoothing (IS) framework introduced in \cite{chauchat2018invariant}. Departing from a system of the form   \eqref{eq:l-is_system}, the goal of smoothing is to find \begin{align}(\chi_i)_i^* = \argmax_{(\chi_i)_i} \mathbb{P}((\chi_i)_i | y_0, \dots, y_n)\label{lik:eq}\end{align}i.e., the maximum a posteriori (MAP) estimate of the trajectory. It is usually found through Gauss-Newton or Levenberg-Marquardt algorithms. First we devise a cost function associated to Problem  \eqref{lik:eq} as the negative log likelihood
	$$
	\mathcal C =-\log\bigl( \mathbb{P}((\chi_i)_{i} | y_0, \dots, y_n)\bigr) 
	$$
	that we seek to minimize. Given a current guess of the trajectory's states, $(\hat{\chi}_i)_i$, the cost function $\mathcal C$  is linearised and then the resulting linear problem is solved exactly, yielding a novel estimate, and so on until convergence. Since $\chi_i$ belongs to a Lie group, linearisation in IS is carried out as
	\begin{equation}
	\forall 0\leq i\leq n\quad \chi_i = \hat{\chi}_i \exp(\xi_i).
	\label{eq:left_inv_error}
	\end{equation}
	where $(\xi_i)_i$ are the searched parameters that minimize the linearized cost. 
    IS linearises the cost $\mathcal C$ as  \cite{chauchat2018invariant}
	\begin{align}
	\tilde {\mathcal  C} =&\|\mathbf{p}_0 + \xi_0\|_{\widetilde{\bfP}_0}^2 
	\label{eq:lie_group_lq}\\
	&+ \sum_i \|\hat{\bfu}_i - \mathbf{F}_i \xi_i + \xi_{i+1}\|_{\mathbf{Q}_i}^2 
	+ \sum_k \|\hat{\bfn}_k + \mathbf{H}_k \Xi\|_{\mathbf{N}_k}^2
	\nonumber
	\end{align}
	where we used the notation $\|\mathbf{Z}\|_{\mathbf{P}}^2=\mathbf{Z}^\top \mathbf{P}^{-1} \mathbf{Z}$, and where $\Xi$ is the concatenation of $(\xi_i)_i$. \eqref{eq:lie_group_lq} relies on the Baker-Campbell-Haussdorff formula \cite{sola2018micro} $\log(\exp(a)\exp(b)) = BCH(a,b)$. $\widetilde{\bfP}_0 = \mathbf{J}_0^{-1} \mathbf{P}_0 \mathbf{J}_0^{-T}$, where $\mathbf{J}_0$ is the left Jacobian of the Lie group $G$ \cite{sola2018micro,chirikjian2011stochastic2}, satisfying $BCH(\mathbf{p}_0, \xi) = \mathbf{p}_0 + \mathbf{J}_0 \xi + o(\|\xi\|^2)$, $\mathbf{p}_0 = \log(\bar{\chi}_0^{-1} \hat{\chi}_0)$ with a prior $\bar{\chi}_0$, $\hat{\bfu}_i = \log(f_i(\hchi_i)^{-1} \hchi_{i+1})$, $\hat{\bfn}_k = \mathbf{y}_k - h_k(\hchi_{I_k})$, and $\mathbf{F}_i, \mathbf{H}_k$ are the (Lie group) Jacobians of $f_i$ and $h_k$ respectively. $\mathbf{H}_k$ was padded with zero blocks for the indices not contained in $I_k$, i.e. not involved in measurement $y_k$.  
	The principle of smoothing algorithms is to solve the linearized problem \eqref{eq:lie_group_lq} in closed form, and to update the trajectory  substituting the optimal $\xi_i$ in \eqref{eq:left_inv_error}. The problem is then relinearised at this new estimate until convergence. 

	\subsection{Group affine Dynamics and Invariant Smoothing}
	In the invariant framework, $f_i$ is assumed to be group affine. These dynamics were introduced in continuous time in  \cite{barrau2017invariant}, and in discrete time in \cite{barrau2019linear}. The main idea is that they extend the notion of linear dynamics (i.e. defined by affine maps)  from vector spaces to Lie groups.
	\begin{definition}Group affine dynamics are defined as
		\begin{align}
		\chi_{i+1} = f_i(\chi_i) = \Lambda_i \phi(\chi_{i})\boUpsilon_i. \label{eq:group_affine_def}
		\end{align}with $\Lambda_i,\boUpsilon_i\in G$, and $\phi$ an automorphism, i.e., satisfies \eqref{auto:prop}.
		\label{def:group_affine}
	\end{definition}
	Group affine dynamics include a large class of systems of engineering interest revolving around navigation and robotics, as shown in e.g. \cite{barrau2017invariant,barrau2019linear, walsli2018invariant, mahony2017geometric}.  
They  come with the \emph{log-linear property}, originally introduced and proved in \cite{barrau2017invariant}.
	
	\begin{prop}[from \cite{barrau2019linear}, discrete-time log-linear  property]\label{longlin:prop}
Let $\chi_{i+1}:=f_i(\chi_i)$ for group affine dynamics \eqref{eq:group_affine_def}. We have \begin{align}f_i(\chi_i \exp(\xi))=\chi_{i+1}\exp(\mathbf{F}_i\xi)\label{loglin:prop2}\end{align} with $\mathbf{F}_i=\bf\mathbf{Ad}_{\boUpsilon_i^{-1}}\mathbf{M}$ a linear operator, and $\mathbf{M}$ from \eqref{eq:LGLA}.
	\end{prop}

    Log-linearity ensures strong properties of invariant smoothing \cite{chauchat2022smoothing}, since dynamics' linearized approximation in \eqref{eq:lie_group_lq} becomes  exact. Moreover, they are easily shown to possess a general preintegration property, see \cite{barrau2019linear, brossard2021associating }, extending that of \cite{forster2016preintegration}. 
    However   the IMU equations are group affine only when sensor biases are neglected,   relying on the Lie group $SE_2(3)$  \cite{barrau2017invariant, barrau2015non, walsli2018invariant}. The recently introduced two-frames group  (TFG) structure partially overcame this limitation, and suggested a new way to account for IMU biases in a principled manner, while unifying most group affine systems discovered so far.


 \section{Localization with an IMU using the TFG}\label{sec3}

This section introduces the  localization problem to which we want to apply invariant smoothing using the TFG. We start by considering the accelerometer biases only. 
 \subsection{Considered simplified problem}
Consider a mobile body  equipped with an inertial measurement unit (IMU) providing gyroscope and accelerometer measurements, and a GNSS (e.g., GPS) receiver providing position measurements $\bfy_i$. We neglect, for now, the gyroscope bias. A simple discretization of the continuous-time equations \cite{forster2016preintegration} yields the  discrete-time dynamics:
 \begin{equation}\label{eq::nav-flat}
\left\lbrace \begin{aligned}
 \bfR_{i+1} & = \bfR_i \exp_m{[\Delta t (\omega_i)_\times]} \\
\bfv_{i+1} & = \bfv_i+\Delta t~(\bfg + \bfR_i \left( \bfa_i - \bfb^a_i \right))\\
\bfp_{i+1} &= \bfp_i + \Delta t ~\bfv_i
\\
\bfb^a_{i+1}  &= \bfb^a_i
\end{aligned} \right. ,
 \end{equation} 
with observation $\bfy_i = \bfp$. In the above $\Delta t$ is a time step, $\bfR_i \in G=SO(3)$ denotes the transformation at time step $i$ that maps the frame attached to the IMU (body) to the earth-fixed  frame, $\bfp_i \in \RR^3$ denotes the position of the body in space, $\bfv_i \in\RR^3$  denotes \color{black}its velocity, $\bfg$ is the earth gravity vector, $\bfa_i,\omega_i \in \RR^3$ the accelerometer and gyroscope signals, $\bfb^a_i$ the accelerometer bias, $\exp_m{}$ denotes the matrix exponential,   and     for any vector $\beta\in\RR^3$, the quantity \color{black} $(\beta)_\times$ denotes the skew-symmetric matrix such that $(\beta)_\times\gamma=\beta\times\gamma$ for any $\gamma\in\RR^3$.


\subsection{Making the system group affine}

To fall into the formalism of Section \ref{sec:lie_group}, one needs to endow the state variables with a Lie group structure. Previously to the TFG theory, it was known that in the absence of IMU biases, inertial navigation is group affine \cite{barrau2017invariant}, but how to deal with the biases was unclear, although there have been recent propositions \cite{fornasier2022equivariant,van_goor2023eqvio}. apart from the latter, most works implementing the invariant framework for inertial navigation - including the authors' -  have treated IMU biases linearly, that is, completed the group structure of $SE_2(3)$, introduced in \cite{barrau2017invariant}, with a linear structure  regarding body variables (biases). 
The considered group composition law thus writes
\begin{align}\label{group:comp2}
\begin{pmatrix} \bfR_1 \\\bfv_1 \\ \bfp_1 \\\bfb^a_1  \end{pmatrix} \biggrouplaw
\begin{pmatrix} \bfR_2  \\\bfv_2\\ \bfp_2\\\bfb^a_2  \end{pmatrix} =
\begin{pmatrix} \bfR_1  \bfR_2 \\\bfv_1 + \bfR_1\bfv_2 \\  \bfp_1 + \bfR_1\bfp_2\\\ \bfb^a_1 +\bfb^a_2\end{pmatrix}\quad\text{(Imperfect IEKF law)}
\end{align}
This group law gave rise to the ``Imperfect IEKF", see \cite{barrau2015non}, leading to ``imperfect invariant smoothing (IS)'' \cite{walsli2018invariant,chauchatPhD}, and led to practical successes \cite{wang2020hybrid,hartley2020contact,wu2017invariant, barrau2018annual,cohen2020navigation}. 
However, none of these approaches, including \cite{fornasier2022equivariant,van_goor2023eqvio},  allowed the biased IMU equations to be group affine.  

In this regard, the two-frames group structure proposed in \cite{barrau2022geometry} was a leap forward. Indeed, it turns out that this structure, which unifies a large number of estimation problems related to navigation, can be cast into the invariant filtering framework using the TFG group law. Following \cite{barrau2022geometry}, the state space can be cast as a two-frame group (TFG). 
The TFG group composition law defines a way to combine the state variables which is defined as follows
\begin{align}\label{group:comp}
\begin{pmatrix} \bfR_1 \\\bfv_1\\ \bfp_1 \\\bfb^a_1  \end{pmatrix} \biggrouplaw
\begin{pmatrix} \bfR_2  \\\bfv_2\\\bfp_2\\\bfb^a_2  \end{pmatrix} =
\begin{pmatrix} \bfR_1  \bfR_2\\\bfv_1 + \bfR_1\bfv_2  \\ \bfp_1 + \bfR_1\bfp_2\\  \bfb^a_2 + \bfR_2^\top\bfb^a_1\end{pmatrix}\quad\text{(TFG law)}
\end{align}
This defines the alternative two-frames group (TFG) law. Its identity element is $(\bfI, \bfzero, \bfzero, \bfzero)$ and the inverse is given by: $
\begin{pmatrix} \bfR ,\bfv, \bfp ,\bfb^a  \end{pmatrix}^{-1}  =
\begin{pmatrix} \bfR^\top ,- \bfR^\top \bfv , - \bfR^\top \bfp , -\bfR \bfb^a\end{pmatrix}.
$

\subsection{Theoretical results}

The TFG structure enables a more principled treatment of sensors' biases, since it leads to group affine properties. 
\begin{prop}[from \cite{barrau2022geometry}]
The IMU equations in 3D with accelerometer bias  are group affine, in the sense of the TFG group structure, whenever
 $  \omega_i = \bfzero$, that is, the orientation of the robot remains unchanged (but it may be arbitrary), while changes in acceleration and velocity are allowed. 
\label{prop:group_affine}
\end{prop}
 \section{Application to biased IMU based localization}
 \label{expl::nav_IMU}

Now that we have recalled the TFG structure, we would like to leverage it for smoothing based localization, which has never been done before. 
In practice, another bias needs to be considered when using an IMU, that of the gyroscope. By accounting for it explicitly, \eqref{eq::nav-flat} then becomes
\begin{equation}\label{eq::nav-flat-gryo-bias}
\left\lbrace \begin{aligned}
 \bfR_{i+1} & = \bfR_i \exp_m{[\Delta t (\omega_i - \bfb_i^\omega)_\times]} \\
\bfv_{i+1} & = \bfv_i + \Delta t~(\bfg + \bfR_i \left( \bfa_i - \bfb^a_i \right))\\
\bfp_{i+1} &= \bfp_i + \Delta t ~\bfv_i
\\
\bfb^a_{i+1}  &= \bfb^a_i,\quad 
\bfb^\omega_{i+1}  = \bfb^\omega_i
\end{aligned} \right. ,
\end{equation}
with $\bfb^\omega_i$ the gyroscope bias. The TFG law includes the latter as follows, where $\chi^{\text{acc}} = \begin{pmatrix} \bfR ,\bfv , \bfp , \bfb^a\end{pmatrix}$:
\begin{align}\label{group:comp_gyro_bias}
\begin{pmatrix} \chi^{\text{acc}}_1 \\\bfb^\omega_1  \end{pmatrix} \biggrouplaw
\begin{pmatrix} \chi^{\text{acc}}_2 \\\bfb^\omega_2  \end{pmatrix} =
\begin{pmatrix} \chi^{\text{acc}}_1 \biggrouplaw \chi^{\text{acc}}_2 \\ \bfb^\omega_2 + \bfR_2^\top\bfb^\omega_1\end{pmatrix}\quad\text{(TFG-gyro law)}
\end{align}
\eqref{eq::nav-flat-gryo-bias} is not group-affine in general since $\exp_m{[\Delta t (\omega_i - \bfb^\omega)_\times]}$ contains state element $\bfb^\omega_i$. This structure can still bring sensible improvements over existing methods \cite{barrau2022geometry}.

\subsection{Computing the Jacobians}
The Jacobian of \eqref{eq::nav-flat} can be directly retrieved from \cite{barrau2022geometry} when $\omega_i = \bfzero$ and there is no gyro bias. It can be extended to the more difficult present setting as follows.
Let $\Omega_i = \exp_m{[\Delta t (\omega_i - \bfb_i^\omega)_\times]}$ and $\bfD_i = D_{\exp_m}(\Delta t (\omega_i - \bfb_i^\omega))$ with $D_{\exp_m}$ the differential of $\exp_m$. Then we have:
    \begin{sizeddisplay}{\footnotesize}
    \begin{equation}
    \bfF_i = \begin{bmatrix}
    \Omega_i^\top - \Delta t ~\bfD_i (\bfb_i^\omega)_\times & & & & - \Delta t ~\bfD_i \\
    - \Omega_i^\top (\Delta t~\bfa_i)_\times & \Omega_i^\top & & -\Delta t~\Omega_i^\top \\
    & \Delta t~\Omega_i^\top & \Omega_i^\top \\
    (\bfb_i^a)_\times \left( \bfI - \Omega_i^\top + \Delta t \bfD_i(\bfb_i^\omega)_\times \right) & & & \bfI \\
    (\bfb_i^\omega)_\times \left( \bfI - \Omega_i^\top + \Delta t \bfD_i(\bfb_i^\omega)_\times \right) & & & & \bfI
    \end{bmatrix}
    \label{eq:full_jac_2TFG}
\end{equation}
\end{sizeddisplay}
It coincides with the Jacobian of ``imperfect IS'' which relies on $SE_2(3) \times \RR^3 \times \RR^3$ (see \cite{brossard2021associating, chauchatPhD}), except for the first block column, whose computation is detailed in Appendix~\ref{app:jacobian}.


\subsection{Differences with other parametrisations}
\label{sec:differences}
Let us summarise the differences between IS based on the TFG structure, imperfect IS which relies on $SE_2(3) \times \RR^3$,  and the smoothing method implemented for NavState in GTSAM \cite{dellaert2016new}, which is a refinement of \cite{forster2016preintegration}. Although the considered residuals and their covariances are essentially the same, the main difference lies in the parametrisation of the state (i.e. the retraction) used to update the state variables at each optimization descent step. Obviously, the treatment of the bias $\bfb = (\bfb^a, \bfb^\omega)$ is a key difference, since all other parametrisations consider it linearly. As concerns the navigational part (attitude, velocity, position), only imperfect IS coincides with IS. More precisely, the retractions used in GTSAM \cite{dellaert2016new} and imperfect IS are respectively
\begin{align}
(\hbfR, \hbfv, \hbfx, \bfb) &\leftarrow (\hbfR \delta_R, \hbfv+\hbfR \delta_v, \hbfx + \hbfR \delta_x, \bfb + \delta_{\bfb}).
\label{eq:gtsam_retraction}\\
(\hbfR, \hbfv, \hbfx, \bfb) &\leftarrow ((\hbfR, \hbfv, \hbfx) \exp_{SE_2(3)}(\delta_R, \delta_v, \delta_x), \bfb + \delta_{\bfb}).
\label{eq:imperfect_retraction}
\end{align}
\eqref{eq:gtsam_retraction} is linear by nature whereas imperfect IS \eqref{eq:imperfect_retraction} uses the exponential map of $SE_2(3)$, which offers a more appropriate nonlinear map. Note that \eqref{eq:gtsam_retraction} is a first-order approximation of \eqref{eq:imperfect_retraction}.  
By constrast, the method proposed herein proposed the full exponential of the TFG, see \eqref{exp:formula} below. 

The Jacobian for GTSAM \eqref{eq:gtsam_retraction} is similar to the one of imperfect IS, replacing $\Omega_i^\top$ and $\bfa_i$ with their estimated counterparts: $\hat{\Omega}_i^\top = \hbfR_{i+1}^\top \hbfR_i$ and $\Delta t~\hbfa_i = \hbfR_i^\top(\hbfv_{i+1} - \hbfv_i - \Delta t~\bfg) - \hbfb_i^a$. Though, it differs more starkly with that proposed herein.

The populated lower-left block of \eqref{eq:full_jac_2TFG} has a slight computational impact on the computations, but only marginal with regard to the whole process.

\section{Experimental Results}
\label{sec:expe}
We evaluate the proposed smoothing method, based on the TFG, on the KITTI dataset \cite{geiger2013kitti}, which contains raw and synchronised IMU data, and a ground-truth.   Smoothing estimates are implemented using sliding windows, where the oldest state is marginalised once a given size is reached. TFG is compared with the two other parametrisations \eqref{eq:gtsam_retraction}, \eqref{eq:imperfect_retraction}, based on a common smoothing implementation to ensure fair comparison. 

When a robot is started, typically a terrestrial wheeled vehicle, it may have no information about its position and its orientation. However, in outdoors applications,
GNSS provides a quick coarse estimate of the position, and the accelerometers that of the vertical direction. The yaw, on the other hand, is much harder to estimate, as no coarse estimate is directly available (magnetometers may be very unreliable due to metallic masses in vehicles and robots). To highlight the differences between the various smoothing methods, we consider the difficult and relevant localization problem where initial  orientation is unknown and thus orientation error may be large ($100~^\circ$ standard deviation), and we focus on the transitory phase during which orientation is recovered (this is called the alignment problem in aerospace engineering, see e.g.,  \cite{wu2013integration, cui2017in_motion, ouyang2022optimization}). 


\subsection{Implementation details}
The IMU data is sampled at 100Hz. We simulate noisy position measurements at 1 Hz by adding white noise to the ground truth with $\sigma_y = 1m$, which is typically the accuracy of a GNSS sensor.

The estimators are compared using three different window sizes of 5, 10 and 15 instants for the state. The initial position is based on the ground truth, roll, pitch, velocity and biases are initialised at zero, and the yaw is randomly sampled.
The initial uncertainty is given by $\sigma_p^0 = 1$m, $\sigma_v^0 = 10$m/s, $\sigma_{R}^0 = 100~^\circ, \sigma_{b^a}^0 = 0.06$m/s\textsuperscript{2}, $\sigma_{b^\omega}^0 = 0.07$rad/s. The process uncertainty uses $\sigma_a = 0.05$m/s\textsuperscript{2}, $\sigma_\omega = 0.01$rad/s, $\sigma_{b^a} = 0.002$m/s\textsuperscript{2}, $\sigma_{b^\omega} = 3.10^{-5}$rad/s. 
For each experiment, 50 Monte Carlo runs are carried out.

\subsection{Results}

When facing large initial errors, optimisation-based estimators easily fall into local minima. In this context, a relevant criterion to compare estimators is to check whether the difficult-to-estimate  yaw error stays inside its believed 3$\sigma$ envelope output by the estimator.   Consistency is thus defined as the ability to convey a consistent (or conservative) estimate of the extent of uncertainty associated with the predictions.

\begin{figure}
    \centering
    \includegraphics[width=0.75\columnwidth]{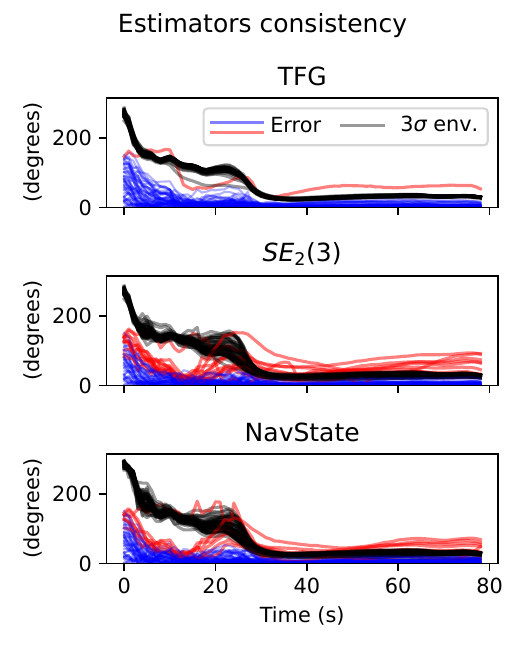}
    \caption{Convergence and consistency comparison of smoothing based on three parametrisations: using the TFG (ours), $SE_2(3)$ \cite{chauchat2022smoothing}, and NavState (GTSAM \cite{dellaert2016new}), with a window size of 5. For 50 Monte Carlo   runs over the KITTI trajectory 01, during the transitory phase after an initialization with random yaw, the yaw error and the $3\sigma$ envelope obtained from the estimated covariance are plotted over time. Consistent trajectories are in blue, inconsistent ones in red.  }
    \label{fig:consistency2}
\end{figure}

Figure  \ref{fig:consistency2} shows the yaw errors over time for the three smoothing methods, highlighting trajectories which yield inconsistent estimates for each of the 50 random initializations, over a typical KITTI trajectory example. We observe a much larger ratio of orientation errors fail to stay inside the 3$\sigma$ envelope (red curves vs blue curves) when using alternative methods to the TFG-based one proposed in the present paper. Besides, we note the 3$\sigma$ envelopes are less dispersed over Monte-Carlo runs when using the TFG, which is consistent with them having less dependency on the estimates, see \cite{barrau2022geometry}.  

To assess the method over a larger number of experiments, Table~\ref{tab:convergence_rate} reports, for the various KITTI trajectories, window sizes and methods, the ratio of Monte Carlo runs which are consistent. As expected, larger windows ensure better convergence (but require more onboard computation capabilities) except when the ratio is already close to one. Moreover, smoothing based on the TFG proves much more robust, in that it systematically outperforms other methods when using the smallest window size (and hence being computationally efficient). It exhibits a special ability to avoid local minima.


\begin{table}[]
    \centering
    \begin{tabular}{|c|c|c|c|c|}
    \hline
       seq.  & window & TFG & $SE_2(3)$ & \eqref{eq:gtsam_retraction} (GTSAM) \\
       \hline
       \multirow{3}{3em}{01} & 5 & \textbf{0.98} & 0.78 & 0.86\\
       & 10 & \textbf{0.98} & \textbf{0.98} & 0.96\\
       & 15 & \textbf{0.98} & \textbf{0.98} & 0.84\\
       \hline
       \multirow{3}{3em}{05} & 5 & \textbf{0.94} & 0.7 & 0.7\\
       & 10 & \textbf{0.96} & 0.92 & 0.92\\
       & 15 & 0.96 & \textbf{0.98} & 0.96 \\
       \hline
       \multirow{3}{3em}{06} & 5 & \textbf{0.96} & 0.68 & 0.66\\
       & 10 & 0.94 & 0.96 & \textbf{1} \\
       & 15 & 0.94 & 0.98 & \textbf{1}\\
       \hline
        \multirow{3}{3em}{07} & 5 & \textbf{0.96} & 0.94 & \textbf{0.96} \\
         & 10 & \textbf{0.98} & \textbf{0.98} & \textbf{0.98}\\
         & 15 & 0.98 & \textbf{1} & \textbf{1} \\
         \hline
        \multirow{3}{3em}{08} & 5 & \textbf{0.98} & 0.84 & 0.86 \\
        & 10 & \textbf{0.98} & \textbf{0.98} & \textbf{0.98} \\
        & 15 & \textbf{1} & 0.98 & 0.94\\
        \hline
        \multirow{3}{3em}{09} & 5 & \textbf{0.84} & 0.68 & 0.7 \\
        & 10 & \textbf{0.94} & 0.86 & 0.88\\
        & 15 & 0.96 & \textbf{1} & 0.96 \\
        \hline
        \multirow{3}{3em}{10} & 5 & \textbf{0.96} & 0.88 & 0.88 \\
        & 10 & \textbf{1} & 0.96 & 0.98\\
        & 15 & \textbf{1} & 0.96 & 0.94\\
        \hline
    \end{tabular}
    \caption{Ratio of consistent trajectories, out of 50 random Monte Carlo initialization, for  various smoothing methods on  the KITTI dataset. Consistency is defined as the yaw error staying inside the 3-$\sigma$ envelope.}
    \label{tab:convergence_rate}
\end{table}

\section{Conclusion}
In this paper we proposed to leverage the recently introduced Two-Frames Group structure in an Invariant Smoothing framework for localization, in a difficult setting where the window size is relatively low, and the inital error may be large.   The necessary derivations, notably to retrieve the Jacobians, helped understanding the properties of the TFG. The proposed method showed increased robustness over the state-of-the-art in terms of convergence on real data. This supports the relevance of the TFG for more complex navigation problems, which opens up for a wide range of future work directions. It may be interesting in the future to investigate notably whether one could  use the framework of GTSAM combined with this new kind of Lie-group based parameterization instead of the current on-manifold parametrization, as to date only improvements have been observed. Further investigations are thus desirable.

\appendices

\section{Computation of \eqref{eq:full_jac_2TFG}}
\label{app:jacobian}
Let $\chi = (R, \bfv, \bfp, \bfb^a, \bfb^\omega)$ be the state. For readability, we denote $\exp(\theta) = \exp_m{[(\theta)_\times]}$ on $SO(3)$, i.e. for $\theta \in \RR^3$. 
Consider an update $\delta_{\chi} = \begin{pmatrix} \delta_{\bfR} & \bfzero & \bfzero & \bfzero & \bfzero \end{pmatrix}$, with $\delta_{\bfR} = \exp(\xi^R)$. Then we have $\chi \bullet \delta_{\chi} = (R \delta_{\bfR}, \bfv, \bfp, \delta_{\bfR}^\top \bfb^a, \delta_{\bfR}^\top \bfb^\omega)$, and
\begin{equation}
    f_i\left(\chi \bullet \delta_{\chi} \right) 
    = \begin{pmatrix} \bfR \delta_{\bfR} \exp{(\Delta t (\omega_i - \delta_{\bfR}^\top~ \bfb^\omega))}\\
    \bfv + \Delta t~(\bfg + \bfR \delta_{\bfR} (\bfa_i - \bfb^a)) \\
    \bfp + \Delta t~\bfv\\
    \delta_{\bfR}^\top~ \bfb^a\\
    \delta_{\bfR}^\top~ \bfb^\omega
    \end{pmatrix}
\end{equation}
Let $\Omega_i = \exp{(\Delta t (\omega_i - \bfb^\omega))}$. Since $\delta_{\bfR}^\top \approx \bfI - (\xi^R)_\times$, we can approximate $\exp{(\Delta t (\omega_i - \delta_{\bfR}^\top~ \bfb^\omega))} \approx \Omega_i \exp{(- \Delta t \bfD_i (\xi^R)_\times \bfb^\omega)}$ with $\bfD_i = D_{\exp}(\Delta t (\omega_i - \bfb^\omega))$ and $D_{\exp}$ the differential of $\exp$ on $SO(3)$. We know that $\bfR \delta_{\bfR} \Omega_i = \bfR \Omega_i \exp{(\Omega_i^\top \xi^R)}$, so $\bfR \delta_{\bfR} \exp{(\Delta t (\omega_i - \delta_{\bfR}^\top~ \bfb^\omega))} \approx \bfR \Omega_i \exp{(\Omega_i^\top \xi^R - \Delta t \bfD_i (\bfb^\omega)_\times \xi^R)}$. Therefore, given \eqref{group:comp_gyro_bias}, we can write
\begin{sizeddisplay}{\small}
\begin{equation}
    f_i\left(\chi \bullet \delta_{\chi} \right) 
    \approx f_i(\chi) \bullet 
    \begin{pmatrix} \exp{(\Omega_i^\top \xi^R - \Delta t \bfD_i (\bfb^\omega)_\times \xi^R)}\\
    - \Delta t \Omega_i^\top (\xi^R)_\times \bfa_i \\
    0\\
    \delta_{\bfR}^\top \bfb^a - \exp{(- \Omega_i^\top \xi^R + \Delta t \bfD_i (\bfb^\omega)_\times \xi^R)} \bfb^a \\
    \delta_{\bfR}^\top \bfb^\omega - \exp{(- \Omega_i^\top \xi^R + \Delta t \bfD_i (\bfb^\omega)_\times \xi^R)} \bfb^\omega
    \end{pmatrix}
    \label{eq:bias_jac_deriv}
\end{equation}
\end{sizeddisplay}
The last terms usually do not appear when considering other parametrisations. Linearising the exponential, we get
\begin{align}
    &\left( \delta_{\bfR}^\top - \exp{(- \Omega_i^\top \xi^R + \Delta t \bfD_i (\bfb^\omega)_\times \xi^R)} \right) \bfb^a \nonumber\\
    &\approx \left( -(\xi^R)_\times + (\Omega^\top \xi^R)_\times - \Delta t (\bfD_i(\bfb^\omega)_\times \xi^R)_\times \right) \bfb^a \nonumber\\
    &= (\bfb^a)_\times \left( \bfI - \Omega^\top + \Delta t \bfD_i(\bfb^\omega)_\times \right) \xi^R
\end{align}
and similarly for $\bfb^\omega$, thus recovering the first block column of \eqref{eq:full_jac_2TFG}. In particular, this highlights why $\Omega_i \neq \bfI$ breaks the group affine property of \eqref{eq::nav-flat} in the 3D case (and without the gyro bias).

\section{Useful Formulas for the TFG}
We recall the results of \cite{barrau2022geometry} being useful herein.
\subsection{Exponential and Logarithm}
The exponential on the considered TFG is given by the following formula:
\begin{align}\label{exp:formula}{
\exp_{TFG}
\begin{pmatrix}
\xi^R \\ \xi^v \\ \xi^p \\ \xi^{b^a} \\ \xi^{b^\omega}
\end{pmatrix}
=
\begin{pmatrix}
\exp(\xi^R) \\
\nu(\xi^R) \xi^v \\ 
\nu(\xi^R) \xi^p \\
\nu(-\xi^R) \xi^{b^a} \\
\nu(-\xi^R) \xi^{b^\omega}
\end{pmatrix}}
\end{align}
Where $\nu$ is given by:
\begin{align*}
\nu(\xi) & = \bfI + \frac{1-\cos(||\xi||)}{||\xi||^2} (\xi)_{\times }+\frac{||\xi||-\sin(||\xi||)}{||\xi||^3}(\xi)_{\times }^2
\end{align*}

Its inverse, the logarithm, writes:
\begin{align}\label{log:formula}{
\log_{TFG}
\begin{pmatrix}
\bfR \\ \bfv \\ \bfp \\ \bfb^a \\ \bfb^\omega
\end{pmatrix}
=
\begin{pmatrix}
\xi^R \\
\nu(\xi^R)^{-1} \bfv \\
\nu(\xi^R)^{-1} \bfp \\
\nu(-\xi^R)^{-1} \bfb^a \\
\nu(-\xi^R)^{-1} \bfb^\omega
\end{pmatrix}}
\end{align}

\subsection{Adjoint Matrices and left Jacobian}
In order to apply \eqref{eq:lie_group_lq} to the TFG, one needs to compute $\widetilde{\bfP}_0 = \mathbf{J}_0^{-1} \mathbf{P}_0 \mathbf{J}_0^{-T}$, where $\mathbf{J}_0$ is the left Jacobian of the group computed at $\bfp_0$, which is given by the following sum $\mathbf{J}_0 = \sum_{j \geq 0} \frac{1}{(n+1)!} (\mathbf{ad_{\bfp_0}})^j$ \cite{sola2018micro}. Moreover $\mathbf{ad_{\bfp_0}}$ can be determined through identification, thanks to the fact that $\exp_m{\mathbf{ad_{\bfp_0}}} = \mathbf{Ad}_{\exp_{TFG} \bfp_0}$. Thus, what is left to do is compute $\mathbf{Ad}_\chi$ for an element of the TFG. In general it is given by the differential of $g \mapsto \chi \bullet g \bullet \chi^{-1}$ around the identity. Let $\chi = (\bfR, \bfv, \bfp, \bfb^a, \bfb^\omega)$ be a state. Then, following \cite{barrau2022geometry}, we get
\begin{equation}
    \mathbf{Ad}_{\chi} = \begin{bmatrix} \bfR \\ (\bfv)_\times \bfR & \bfR \\ (\bfp)_\times \bfR & & \bfR \\
    \bfR (\bfb^a)_\times & & & \bfR \\
    \bfR (\bfb^\omega)_\times & & & & \bfR \end{bmatrix} 
    \label{eq:Ad_chi}
\end{equation}
Now, let $\chi = \exp_{TFG}(\xi)$, with $\xi = (\xi^R, \xi^p, \xi^p, \xi^{b^a}, \xi^{b^\omega})$. Using \eqref{exp:formula}, and the facts that $\bfR \nu(-\xi^R) = \nu(\xi^R)$, and $\bfR (\bfx)_\times = (\bfR \bfx)_\times \bfR$, we get
\begin{equation}
    \mathbf{Ad}_{\exp_{TFG}(\xi)} = \begin{bmatrix} \bfR \\ 
    (\nu(\xi^R) \xi^v)_\times \bfR & \bfR \\
    (\nu(\xi^R) \xi^p)_\times \bfR & & \bfR \\
    (\nu(\xi^R) \xi^{b^a})_\times \bfR & & & \bfR \\
    (\nu(\xi^R) \xi^{b^\omega})_\times \bfR & & & & \bfR \end{bmatrix} 
    \label{eq:Ad_exp}
\end{equation}
Therefore, we can directly identify
\begin{equation}
    \mathbf{ad}_\xi = \begin{bmatrix} (\xi^R)_\times \\ 
    (\xi^v)_\times & (\xi^R)_\times \\ (\xi^p)_\times & & (\xi^R)_\times \\
    (\xi^{b^a})_\times & & & (\xi^R)_\times \\
    (\xi^{b^\omega})_\times & & & & (\xi^R)_\times \end{bmatrix}
\end{equation}

\bibliography{biblio}
\bibliographystyle{plain}

\end{document}